\documentclass{article}

\usepackage[final,nonatbib]{nips_2016}

\usepackage{algorithm}
\usepackage[utf8]{inputenc} 
\usepackage[T1]{fontenc}    
\usepackage{hyperref}       
\usepackage{url}            
\usepackage{booktabs}       
\usepackage{amsfonts}       
\usepackage{nicefrac}       
\usepackage{microtype}      
\usepackage{amsmath}       
\usepackage{amsthm}
\usepackage{gensymb}
\usepackage{subcaption}
\usepackage{bm}
\usepackage{amssymb}

\usepackage{framed}
\usepackage{tikz}
\usepackage{float}
\usepackage{multirow}
\usepackage{todonotes}
\usepackage{changebar}
\usepackage{soul}
\usepackage{color}
\usepackage{authblk}
\usepackage{algorithm}

\usepackage[english]{babel}
\usepackage[utf8]{inputenc}
\usepackage{algorithm}
\usepackage[noend]{algpseudocode}

\title{Mixed Low-precision Deep Learning Inference using Dynamic Fixed Point}

\author[1]{Naveen Mellempudi}
\author[1]{Abhisek Kundu}
\author[1]{Dipankar Das}
\author[1]{Dheevatsa Mudigere}
\author[1]{Bharat Kaul}
\affil[1]{Parallel Computing Lab, Intel Labs \authorcr Bangalore, India}

\newcommand{\FsNorm }[1]{\mbox{}\|#1\|_F  }

\newcommand{\setlinespacing}[1]%
           {\setlength{\baselineskip}{#1 \defbaselineskip}}

\newcommand{\abs }[1]{\left|#1\right|}
\newcommand{\sabs }[1]{|#1|}

\newcommand{\mat}[1]{{\ensuremath{\bm{\mathrm{#1}}}}}

\def\matW{\mat{W}}

\def\frac#1#2{{#1\over #2}}

\def\floor#1{{\left\lfloor\,#1\,\right\rfloor}}

\def\dotfil{\leaders\hbox to 1.5mm{.}\hfill}
\newcounter{rmnum}
\def\RN#1{\setcounter{rmnum}{#1}\uppercase\expandafter{\romannumeral\value{rmnum}}}
\def\rn#1{\setcounter{rmnum}{#1}\expandafter{\romannumeral\value{rmnum}}}

\begin{document}

\maketitle

\begin{abstract}
We propose a cluster-based quantization method to convert pre-trained full precision weights into ternary weights with minimal impact on the accuracy. In addition we also constrain the activations to 8-bits thus enabling sub 8-bit full integer inference pipeline. Our method uses smaller clusters of N filters with a common scaling factor to minimize the quantization loss, while also maximizing the number of ternary  operations. We show that with cluster size of N=4 on Resnet-101, can achieve $71.8\%$ TOP-1 accuracy, within $6\%$ of the best full precision result, while replacing $\approx85\%$ of all multiplications with 8-bit accumulations. Using the same method with 4-bit weights achieves $76.3\%$ TOP-1 accuracy which within $2\%$ of the full precision result. We also study the impact of the size of the cluster on both performance and accuracy, larger cluster sizes N=64 can replace $\approx98\%$ of the multiplications with ternary operations but introduces significant drop in accuracy which necessitates fine tuning the parameters with retraining the network at lower precision. To address this we have also trained low-precision Resnet-50 with 8-bit activations and ternary weights by pre-initializing the network with full precision weights and achieve $68.9\%$ TOP-1 accuracy within 4 additional epochs. Our final quantized model can run on a full 8-bit compute pipeline, with a potential 16x improvement in performance compared to baseline full-precision models.

\end{abstract}

\section{Introduction}

Deep Learning has achieved unparalleled success with large-scale machine learning. Deep Learning models are used for achieving state-of-the-art results on a wide variety of tasks including Computer Vision, Natural Language Processing, Automatic Speech Recognition and Reinforcement Learning \cite{2016DlBook}. Mathematically this involves solving a complex non-convex optimization problem with order of millions or more parameters. Solving this optimization problem - also referred to as training the neural network is a compute-intensive process that for current state-of-art networks requires days to weeks. Once trained, the DNN is used by evaluating this many-parameter function on specific input data - usually referred to as inference. While the compute intensity for inference is much lower than that of training, inference also involves significant amount of compute. Moreover, owing to the fact that inference is done on a large number of input data, the total computing resources spent on inference is likely to dwarf those that are spent on training. Due to the large and somewhat unique compute requirements for both deep learning training and inference operations, it motivates the use of non-standard customized arithmetic \cite{hubara2016qnn,courbariaux2016bnn,hubara2016bnn,zhou2016dorefa,logqunat,2016twn} and specialized compute hardware to run these computations as efficiently as possible \cite{gupta2015lp,zhu2016ttq,venkatesh2016,finn2016}. Furthermore, there some theoretical evidence and numerous empirical observations that deep learning operations can be successfully done with much lower precision.

In this work we focus on reducing the compute requirements for deep learning inference, by directly quantizing pre-trained models with minimum (or no) retraining and achieve near state-of-art accuracy.\\
\\Our paper makes the following contributions:
\begin{enumerate}
\item We propose a novel cluster-based quantization method to convert pre-trained weights to lower precision representation with minimal loss in test accuracy. 
\item On Resnet-101 with 8-bit activations and using cluster size (N=4) to quantize weights, we achieve $76.3\%$ TOP-1 accuracy with 4-bit weights and $71.8\%$ TOP-1 accuracy with 2-bit ternary weights. To the best of our knowledge this is the best reported accuracy with ternary weights on ImageNet dataset\cite{imagenet_data}, without retraining the network.
\item We explore the performance-accuracy trade-off using different cluster sizes with ternary weight representation. For a cluster size of N, we reduce the higher precision ops (8-bit multiply) to one for every $N*K^2$ lower precision ops (8-bit accumulation), which results significant reduction in computation complexity. 
Using smaller cluster size of N=4 we achieve state-of-the-accuracy, but larger cluster sizes (N=64) would require retraining the network at lower precision to achieve comparable accuracy.
\item We train a pre-initialized low precision Resnet-50 using 8-bit activations and 2-bit weights using larger cluster (N=64) and achieve $68.9\%$ TOP-1 accuracy on ImageNet dataset\cite{imagenet_data} within $4$-epochs of fine-tuning.  
\end{enumerate}

\section{Related Work}

Deep learning training and inferencing are highly compute intensive operations, however using full precision $(FP32)$ computations on conventional hardware is inefficient and not strictly warranted from functional point-of-view. To address this issue, there has been a lot of interest at using lower precision for deep learning, in an attempt to identify the minimum required precision to ensure functional correctness within acceptable thresholds. 

In the past many researches have proposed low-precision alternatives to perform deep learning tasks. Vanhoucke et al.\cite{vanhoucke2011improving} showed that using 8-bit fixed-point arithmetic convolution networks can be sped up by up to 10x on speech recognition tasks on general purpose CPU hardware. Gupta et al.\cite{gupta2015lp} have successfully trained networks using 16-bit fixed point on custom hardware. Miyashita et al.\cite{logqunat} used log quantization on pre-trained models and achieved good accuracy by tuning the bit length for each layer. More recently, Venkatesh et al.\cite{venkatesh2016} achieved near state of the art results using 32b activations with 2-bit ternary weights on Imagenet dataset. Hubara et al.\cite{hubara2016bnn} have demonstrated that with weights as binary values training from scratch can achieve near state-of-the-art results for ILSVRC 2012 image classification task\cite{imagenet_contest}.
\section{Low Precision Inference}{\label{sec_lpinf}}
In this paper, we primarily focus on improving the performance and accuracy of the inference task. We explore possibility of achieving high accuracy using sub 8-bit precision on state-of-the-art networks without expensive retraining. Previous work from Miyashita et al.\cite{logqunat} showed that by compressing the dynamic range of the input, it is possible to minimize the quantization loss and achieve high accuracy. 
We take a different approach to minimize the impact of dynamic range on quantization. We propose a cluster-based quantization method that groups weights into smaller clusters and quantize each cluster with a unique scaling factor. We use static clustering to group filters that accumulate to the same output feature to simplify the convolution operations. Empirical evidence also suggests that these clusters which learn similar features tend to have smaller dynamic range. Using dynamic fixed point representation, this method can effectively minimize the quantization errors and improve the inference accuracy of quantized networks. Applying this scheme on a pre-trained Resnet-101 model, with $4$-bit weights and $8$-bit activations, we achieve $76.3\%$ TOP-1 accuracy on ImageNet dataset\cite{imagenet_data}, without any retraining.

\subsection{2-bit Ternary Weights}
{\label{sec_ternary}}
Going below 4-bits we use the the ternary representation for weights, following the threshold based approximation proposed by Li et al \cite{2016twn}, i.e., approximate full-precision weight $\matW \approx \alpha\hat\matW$ in $\ell_2$ norm $\FsNorm{\cdot}$, where $\alpha$ is a scaling factor and $\hat\matW$ is a ternary weight with $\hat\matW_i \in \{-1,0,+1\}$. Where $\matW$ is the matrix representing learned full-precision weights, and $\hat\matW$ represents the corresponding ternary representation. We apply the block-quantization method described in section-\ref{sec_lpinf}, to compute multiple scaling factors for each layer to minimize the accuracy loss.  
Our method differs from \cite{2016twn} in the approximation used for computing scaling factor ($\alpha$). We use the $RMS$ formulation as shown by the equation (\ref{eq_1}). The intuition behind using $RMS$ term is to push the threshold parameter towards larger values within the cluster which helps speed up weight pruning.

\begin{equation}\label{eq_1}
\alpha = \sqrt[]{\dfrac{{\sum_{i\in I_\tau} \matW_i ^2}}{\abs{I_\tau}}}, \text{where } \abs{I_\tau} \text{ is the number of elements in } I_\tau. 
\end{equation}

In addition, we run our search algorithm\ref{alg:TwoR} in hierarchical fashion by minimizing the error within each filter first and then within the cluster of filters. Experimental evidence shows that these improvements help finding the optimal scaling factor that minimizes quantization loss.

Using multiple scaling factors can lead to more 8-bit multiplications. Hence, we choose the cluster size carefully to improve the ratio of low-precision (2-bit) to high-precision (8-bit) operations (Section \ref{sec_perfimpl}).
Our algorithm (Algorithm \ref{alg:TwoR}) takes the full-precision learned weights and returns clusters of ternary representation of groups of kernels along with their scaling factors. We further quantize the scaling factors down to 8-bit to eliminate any operation that requires more than 8 bits. Applying this scheme on pre-trained ResNet-101 model, using 8-bit activations we achieve $71.8\%$ TOP-1 accuracy on ImageNet dataset.

\begin{algorithm}[t]
\centerline{
\caption{Ternarize Weights}\label{alg:TwoR}
}
\begin{algorithmic}[1]
\State \textbf{Input:} Learned full-precision weights $\matW$ of a layer with $d$ filters.
%
\State Group filters into $k$ clusters: $\{G_j\}, j=1,..., k$. 
Let $N = \sabs{G_j}$ (number of filters in $G_j$).
\State For each cluster $G_j$
\State \quad Run Algorithm \ref{alg:threshold} on each filter $\matW \in G_j$, and store the thresholds as a vector $\alpha$.
\State \quad For $t=1,..., N$, \quad 
$T_t=\{i: \alpha_i \text{ belongs to the top } t \text{ elements of sorted } \alpha\}$.
\State \qquad Set $\alpha_t  = \sqrt{\sum_{i\in T_t}\alpha_i^2/\sabs{T_t}}$. 
\State \qquad Construct $\hat\matW{}^{(t)}$, such that, $\hat\matW{}^{(t)}_i = Sign(\matW_i)$, if $\sabs{\matW_i} > \alpha_t$, and 0 otherwise.
\State \quad Find $\alpha_{t^*}$ and $\hat\matW{}^*{}^{(t)}$ that minimizes 
$\sum_{\matW\in G_j}\FsNorm{\matW-\alpha_t\hat\matW{}^{(t)}}^2$.
\State \quad Let $\hat\alpha_{t^*}$ be a reduced-precision representation of $\alpha_{t^*}$.
\State \textbf{Output:} $k$ number of $\hat\alpha_{t^*}$ and the group of ternary weights $\hat\matW$.
\end{algorithmic}
\end{algorithm} 

\begin{algorithm}[t]
\centerline{
\caption{Threshold Selection}\label{alg:threshold}
}
\begin{algorithmic}[1]
\State \textbf{Input:} $\matW \in \mathbb{R}^n$.
%
\State Sort elements of $\matW$ according to magnitude.
\State For $\tau \in [0,1]$, $I_\tau=\{i: \sabs{\matW_i} \text{ belongs to the top } \floor{\tau\cdot n} \text{ elements of sorted list}\}$.
\State Construct $\hat\matW{}^{(\tau)}$, such that, $\hat\matW{}^{(\tau)}_i = Sign(\matW_i)$, for $i \in I_\tau$, and 0 otherwise.
\State Set $\alpha_\tau = \sqrt{\sum_{i\in I_\tau} \matW_i ^2/\sabs{I_\tau}}$.
\State Compute $\alpha_{\tau^*}$ that minimizes 
$\FsNorm{\matW-\alpha_\tau\hat\matW{}^{(\tau)}}^2$, for $\tau \in [0,1]$.
\State \textbf{Output:} $\alpha_{\tau^*}$
\end{algorithmic}
\end{algorithm}




\subsection{C1 and BatchNorm Layers} 
In our experiments we keep weights of the first convolution layers at $8$-bits to prevent from accumulating losses while the rest of the layers including fully connected layers operate at lower precision. We also recompute the batch norm parameters during the inference phase to compensate for the shift in variance introduced by quantization. This is essential for making it work, when we are not retraining at lower precision. We are exploring the possibility of fusing batch normalization layers with the convolution layers before quantization to avoid this extra computation. 

\begin{figure}[h]
\centering
\includegraphics{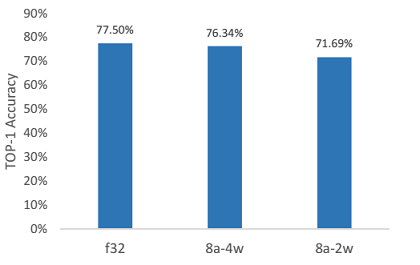}
\centering
\caption{Resnet-101 results on ImageNet dataset using 8-bit activations with 4-bit weights $8a-4w$ and 2-bit weights $8a-2w$} \label{fig_res101}
\end{figure}

\subsection{Performance Implications}
{\label{sec_perfimpl}}
Choosing the right cluster size is a trade-off between  performance and accuracy, while having one cluster per layer favors higher compute density by eliminating all multiplications, it's not ideal for achieving high accuracy. Although, previous research in this space\cite{venkatesh2016} showed that it is possible to recover some of this lost accuracy through retraining. It's not always ideal, because of the costs involved in retraining these networks in low-precision, not to mention the technical difficulties involved in achieving reasonable solution on these networks. 

We explored the accuracy-performance trade-off with various cluster sizes. Our experiments show on Resnet-101, using a cluster size of $N=4$ can achieve $71.8\%$ TOP-1 accuracy, within $6\%$ of the full precision result. This result significant because this is to the best of our knowledge highest accuracy achieved on Imagenet dataset\cite{imagenet_data} without retraining the network in low-precision. In terms of performance impact, the clustering will result in one 8-bit multiplication for the entire cluster ($N*K^2$) of ternary accumulations. Assuming roughly $50\%$ of the convolutions are 3x3 and the rest are 1x1, using this block size can potentially replace $85\%$ of multiplications in Resnet-101 convolution layers with simple $8$-bit accumulations. For networks that predominantly use filters that are 3x3 or bigger, this ratio would be greater than $95\%$. We explored the accuracy-performance trade-off with various cluster sizes, we concluded that using cluster size of $N=64$, we can replace $\approx98\%$ of multiplications in Resnet-101 with $8$-bit accumulations, but with a significant loss to the accuracy. At this point retraining the network at lower precision would be necessary.


\section{Training with Low-precision}
{\label{sec_lptrain}}
We trained the low precision ResNet-50 on ImageNet dataset using 2-bit weights and 8-bit activations by initializing the network with pre-trained full precision model. We take the approach proposed by Marcel et al.\cite{marcelbatchnorm}, and replace data pre-processing steps such as mean-subtraction and jittering with batch normalization layer inserted right after the data later. We obtained the pre-trained models published by Marcel et al.\cite{marcelbatchnorm} and fine-tune the parameters of our low-precision network. In the forward pass, the weights are converted to 2-bit ternary values using the algorithm described in \ref{alg:TwoR} in all convolution layers, except the first layer, where the weights are quantized to 8-bit fixed point representation. Activations are quantized to 8-bit fixed point in all layers including ReLU, BatchNorm layers. We did not quantize the weights in FC layer for the training exercise. Gradient updates are performed in full precision for convolution and FC layers. We reduced the learning rate to an order of 1e-4, in order to avoid exploding gradients problem, while we keep all the other hyper parameters same as that of full precision training. After running for 4-epochs, we recovered most of the accuracy and achieved $68.6\%$ Top-1 and $88.7\%$ Top-5 accuracy compared to our baseline $75.02\%$(Top-1), $92.2\%$(Top-5). 

\begin{figure}[h]
\centering
\includegraphics{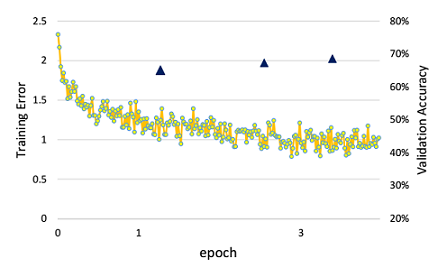}
\centering
\caption{Fine-tuning Resnet-50 with pre-initialized weights on Imagenet dataset.} \label{fig_res101}
\end{figure}


\section {Conclusion}
We propose a clustering based quantization method which exploits local correlations in dynamic range of the parameters to minimize the impact of quantization on overall accuracy. We demonstrate near SOTA accuracy on Imagenet data-set using pre-trained models with quantized networks without any low precision training. On Resnet-101 using 8-bit activations the error from the best published full precision $(FP32)$ result is within $\approx6\%$ for ternary weights and within $\approx2\%$ for 4-bit weights. To the best of our knowledge this is the best achieved accuracy with ternary weights for Imagenet dataset.

Our clustering based approach allows for tailoring solutions for specific hardware, based on the accuracy and performance requirements. Smaller cluster sizes  achieves best accuracy, with N=4 $\approx85\%$ of the computations as low precision operations (simple 8-bit accumulations) and this is better suited for implementation on specialized hardware. Larger cluster sizes are more suited to current general purpose hardware, with a larger portion of computations as low precision operations ($>98\%$ for N=64), however this comes with the cost of reduced accuracy. This gap can be bridged with additional low precision training as show in section \ref{sec_lptrain}, work is underway to further improve this accuracy. Our final quantized model can be efficiently run on full 8-bit compute pipeline, thus offering a potential $16X$ performance-power benefit.

Furthermore as continuation of this work, we are looking in a more theoretical exploration to better understand the formal relationship between the clustering and final accuracy, with an attempt establish realistic bounds for given network-performance-accuracy requirement.


\newpage
\bibliographystyle{plain}
\bibliography{references}

\end{document}